\documentclass{article}

\usepackage{microtype}
\usepackage{graphicx}
\usepackage{subfigure}
\usepackage{booktabs} 
\usepackage[table,xcdraw]{xcolor}
\usepackage{hyperref}
\usepackage{amsmath}
\usepackage{amssymb}

\renewcommand{\vec}[1]{\mathbf{\boldsymbol{#1}}}

\newcommand{\kvec}{\vec{k}}

\newcommand{\qvec}{\vec{q}}

\newcommand{\vvec}{\vec{v}}

\newcommand{\xvec}{\vec{x}}
\newcommand{\yvec}{\vec{y}}
\newcommand{\zvec}{\vec{z}}

\usepackage[accepted]{icml2021}
\usepackage{amsmath,amssymb,enumerate,bbm,color,alltt}

\icmltitlerunning{
Poolingformer: Long Document Modeling with Pooling Attention
}

\begin{document}

\twocolumn[
\icmltitle{Poolingformer: Long Document Modeling with Pooling Attention}



\icmlsetsymbol{equal}{*}

\begin{icmlauthorlist}
\icmlauthor{Hang Zhang}{SCU,Intern}
\icmlauthor{Yeyun Gong}{MSRA}
\icmlauthor{Yelong Shen}{MSD365}
\icmlauthor{Weisheng Li}{USTC}
\icmlauthor{Nan Duan}{MSRA}
\icmlauthor{Weizhu Chen}{MSD365}
\icmlauthor{Jiancheng Lv}{SCU}
\end{icmlauthorlist}

\icmlaffiliation{SCU}{College of Computer Science, Sichuan University}
\icmlaffiliation{Intern}{During Internship at MSRA}  
\icmlaffiliation{MSRA}{Microsoft Research Asia}
\icmlaffiliation{MSD365}{Microsoft Azure AI}
\icmlaffiliation{USTC}{University of Science and Technology of China}

\icmlcorrespondingauthor{Yeyun Gong}{yegong@microsoft.com}
\icmlcorrespondingauthor{Weizhu Chen}{wzchen@microsoft.com}
\icmlkeywords{Long Document Modeling, Machine Learning, ICML}

\vskip 0.3in
]



\printAffiliationsAndNotice{}  

\begin{abstract}


In this paper, we introduce a two-level attention schema, Poolingformer, for long document modeling. Its first level uses a smaller sliding window pattern to aggregate information from neighbors. Its second level employs a larger window to increase receptive fields with pooling attention to reduce both computational cost and memory consumption. We first evaluate Poolingformer on two long sequence QA tasks: the monolingual NQ and the multilingual TyDi QA. Experimental results show that Poolingformer sits atop three official leaderboards measured by F1, outperforming previous state-of-the-art models by 1.9 points (79.8 vs. 77.9) on NQ long answer, 1.9 points (79.5 vs. 77.6) on TyDi QA passage answer, and 1.6 points (67.6 vs. 66.0) on TyDi QA minimal answer. We further evaluate Poolingformer on a long sequence summarization task. Experimental results on the arXiv benchmark continue to demonstrate its superior performance.  
\end{abstract}
\section{Introduction}
\label{Introduction}

Transformer~\cite{vaswani2017attention} architecture has been widely used in various natural language processing tasks with impressive results such as Translation~\cite{lewis2020bart}, Summarization~\cite{prophetnet}, Text Classification~\cite{he2020deberta} , and Language Modeling~\cite{BrownMRSKDNSSAA20}. Self-attention is one of the key components in Transformer, which allows text tokens to interact with each other, and produce contextual representations. Despite the effectiveness of self-attention, its computational and memory complexity increases quadratically with respect to the sequence length. Therefore, most of existing transformer-based pretrained models ~\cite{alberti2019bert,he2020deberta,liu2019roberta} set the maximum sequence length to 512 due to either memory or computational constraints, which often leads to a worse performance in long sequence  tasks \cite{kwiatkowski2019natural, cohan2018discourse}.

\begin{figure}[t]  
    \centering
    \subfigure[Single-level local attention]{
        \centering
        \includegraphics[width = 0.45\linewidth]{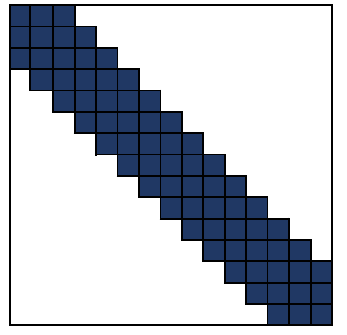}
        }
    \subfigure[Two-level pooling attention]{
        \centering
        \includegraphics[width = 0.45\linewidth]{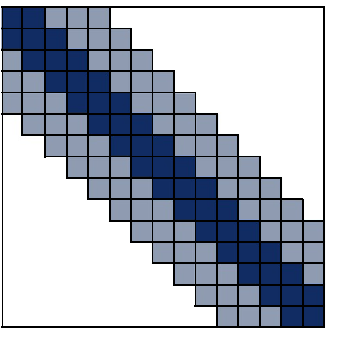}
        }
\caption{(a): The receptive field of single-level local attention  (b): The receptive field of our two-level pooling attention.}
    \label{figure.visonofwindow}
\end{figure}
A lot of works have been proposed to adapt the self-attention layer in transformer to better model long sequence  \cite{miculicich2018-document,liu2019hierarchical, beltagy2020longformer,bigbird,wang2020cluster}. 
For example, Longformer ~\cite{beltagy2020longformer} puts forward a combination of both local and global attention patterns to reduce computational cost. Hierarchical transformer ~\cite{liu2019hierarchical} proposes to split the long document into shorter paragraphs, and apply inter-self-attentions within a paragraph and intra-self-attentions across paragraphs. 


Inspired by previous works \cite{miculicich2018-document,liu2019hierarchical, beltagy2020longformer,bigbird,wang2020cluster}, we propose Poolingformer in which it revises the full self-attention to a two-level attention schema. The first level adopts a sliding-window attention pattern where each token only attends to its neighbor tokens within the window, as shown in Figure~\ref{figure.visonofwindow} (a).  In the second level attention, it increases the receptive fields with a larger window size, followed by a pooling operation on both the key and value vectors in transformer to decrease the number of tokens to be attended. This multi-level design combining both sliding window and pooling can significantly reduce the computational cost and memory consumption while still attain exceptional model performance. Compared with the models with single-level local attention~\cite{beltagy2020longformer,bigbird}, Poolingformer allows a larger attention receptive field per token via the benefit from the second-level pooling attention mechanism, as shown in Figure~\ref{figure.visonofwindow} (b). In the meantime, it preserves the sliding-window attention pattern at the first-level to mitigate the information loss due to the pooling operation at the second-level attention. Compared with Hierarchical Transformer ~\cite{liu2019hierarchical}, Poolingformer obviates the need to explicitly split a long document into paragraphs. Thus, it is a more general long-sequence model which can be applied to extremely long text sequence in a cohesive manner. Compared with Transformer~\cite{vaswani2017attention}, the computational and memory complexity of Poolingformer only increase linearly with respect to sequence length. 

In the experiment, we first demonstrate the superior performance of Poolingformer using two QA datasets: the monolingual NQ\footnote{\url{https://ai.google.com/research/NaturalQuestions/dataset}} and multilingual TyDi QA\footnote{\url{https://ai.google.com/research/tydiqa}}. Experimental results show that Poolingformer has achieved new state-of-the-art results on their official leaderboards. We continue to evaluate Poolingformer on the extremely long summarization task arXiv~\cite{cohan2018discourse}. Experimental results show Poolingformer has set up new state-of-the-art results on this challenging benchmark.

\section{Model}
In the section, we present the model architecture of Poolingformer. We start with an introduction to the self-attention mechanism in Transformer model in section \ref{sec:self_att} and elaborate the details of Poolingformer self-attention in section~\ref{sec:poolingfomer_att}. 
\subsection{Transformer Self-Attention}
\label{sec:self_att}
Given a sequence of text embeddings denoted as ${\mathbf{X}} = ({\xvec_1},  {\xvec_2}, ..., {\xvec_n})$, $n$ is the text sequence length and ${\xvec_i} \in  \mathbb{R}^d$ is the embedding vector of the $i$-th token. In the transformer model, it produces the query, key, and value vectors for each token by a linear projection of the embeddings, as in Eqn \ref{eq:qkv}. 
\vspace{-1.2mm}
\begin{align}
 \begin{pmatrix}
    {\mathbf{Q}}\\
    {\mathbf{K}}\\
    {\mathbf{V}}\\
  \end{pmatrix}= \begin{pmatrix}
    {\mathbf{W_q}}\\
    {\mathbf{W_k}}\\
    {\mathbf{W_v}}\\  
  \end{pmatrix}{\mathbf{X}} +
\begin{pmatrix}
    {\mathbf{b_q}}\\
    {\mathbf{b_k}}\\
    {\mathbf{b_v}}\\  
  \end{pmatrix}
 \label{eq:qkv}
 \end{align}
where $\mathbf{Q}$, $\mathbf{K}$ and $\mathbf{V}$ are the query, key and value matrices respectively. Specifically, let $\qvec_i$ be the $i$-th column of matrix $\mathbf{Q}$ which indicates the $i$-th token's query vector, $\kvec_i$ and $\vvec_i$ are defined in the same way.   

A typical self-attention mechanism computes the inner product between the query and key vectors as the attention scores, and performs weighted intra-aggregation of  value vectors to produce contextualized representations. For instance, token $i$'s output vector $\yvec_i$ is calculated in Eqn.~\ref{eq:attention}.     
\begin{align}
\yvec_i^T = \text{Softmax}\left( \alpha \qvec_i^T \mathbf{K} \right) \mathbf{V}^T
\label{eq:attention}
\end{align}
where $\alpha$ is a constant scalar and usually set as: $\alpha=1/\sqrt{d}$.  
Therefore, the computation of the full self-attention comes with a ${O}(n ^2)$ memory and computational complexity, which limits its ability for processing extremely long text sequence.



\begin{figure*}[ht] 
    \centering
    \includegraphics[width=0.8\linewidth]{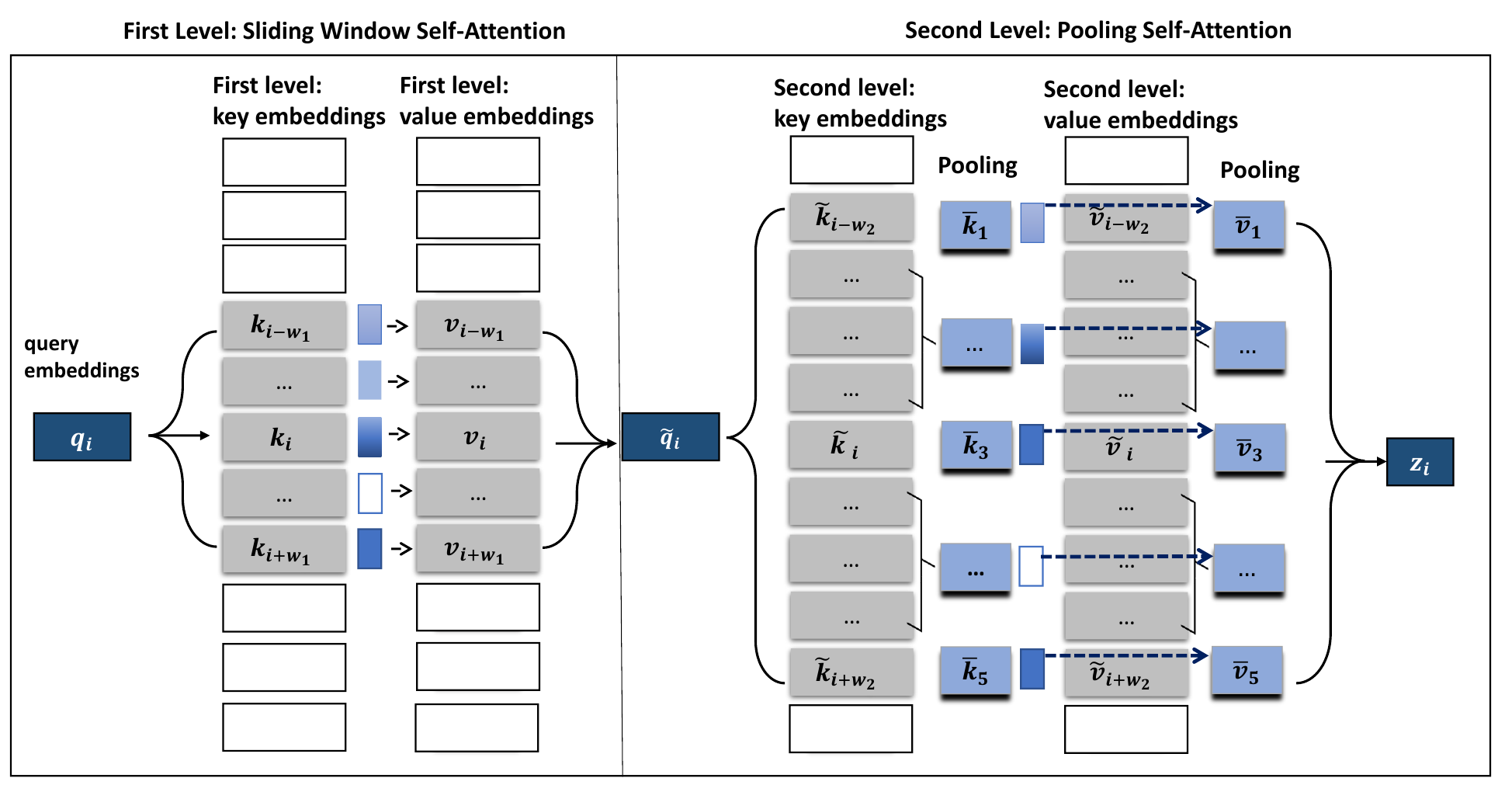}
    \caption{The illustration of the two-level self-attention in PoolingFormer. Left block is the first level sliding window attention; Right block is the second level pooling attention.} \label{fig:arch_poolformer}
\end{figure*}
\subsection{Poolingformer Self-Attention}

\label{sec:poolingfomer_att}
Poolingformer revises the full self-attention mechanism to a two-level attention schema: the first level attention adopts the sliding window pattern to let each token only attend to its neighbor tokens within the window; the second level attention increases the receptive fields with a larger window size and performs attention over pooled key and value matrices. We provide an illustration of the Poolingformer self-attention in Figure \ref{fig:arch_poolformer} with more details elaborated in following subsections.


\subsubsection{First-Level: Sliding Window Attention}

The first level self-attention sets a sliding window attention pattern to allow each token only attend to its neighbor tokens. For instance, token $i$'s neighbor set within the window size $w_1$ is defined as $\mathcal{N}{(i,w_1)}$:
\begin{equation}
    \mathcal{N}{(i,w_1)} = \left\{ i - w_1, ..., i, ..., i + w_1  \right\}
\label{eq.receptivefield}
\end{equation}
The sub-matrices of $\mathbf{K}$ and $\mathbf{V}$ with corresponding column indexes $\mathcal{N}{(i,w_1)}$ are denoted by $\mathbf{K}_{\mathcal{N}{(i,w_1)}}$ and $\mathbf{V}_{\mathcal{N}{(i,w_1)}}$. According to the sliding window pattern, each $\qvec_i$ only attends to neighbor set $\mathcal{N}{(i,w_1)}$. Therefore, token $i$'s output of the first-level attention is computed as:
\begin{align}
\yvec_i^T = \text{Softmax}\left( \alpha \qvec_i^T \mathbf{K}_{\mathcal{N}{(i,w_1)}} \right) \mathbf{V}^T_{\mathcal{N}{(i,w_1)}}
\label{eq:first_att}
\end{align}


Since size of the receptive field is limited to $w_1$, it could lead to a worse model performance for long document understanding tasks. 


\subsubsection{Second-Level: Pooling Attention}
\label{subsubsection:poolingattention}
The second level pooling attention module is built upon the outputs of the first level attention $\mathbf{Y} = (\yvec_1, ..., \yvec_n)$. 
It first produces new query, key and value matrices from $\mathbf{Y}$:
\vspace{-1.2mm}
\begin{align}
 \begin{pmatrix}
    {\mathbf{\widetilde{Q}}}\\
    {\mathbf{\widetilde{K}}}\\
    {\mathbf{\widetilde{V}}}\\
  \end{pmatrix}= \begin{pmatrix}
    {\mathbf{W_{\widetilde{q}}}}\\
    {\mathbf{W_{\widetilde{k}}}}\\
    {\mathbf{W_{\widetilde{v}}}}\\
  \end{pmatrix}{\mathbf{Y}} +
\begin{pmatrix}
    {\mathbf{b_{\widetilde{q}}}}\\
    {\mathbf{b_{\widetilde{k}}}}\\
    {\mathbf{b_{\widetilde{v}}}}\\  
  \end{pmatrix}
 \label{eq:qkv_2}
 \end{align}
The query vector of token $i$ and its corresponding key/value matrices are $\widetilde{\qvec}_i$, $\widetilde{\mathbf{{K}}}_{\mathcal{N}(i, w_2)}$, and $\widetilde{\mathbf{{V}}}_{\mathcal{N}(i, w_2)}$ respectively, with a larger window size $w_2$. ($w_2$ can be set to $n$ in the extreme case). Since $w_2$ could be very large, we apply a pooling layer to compress $\widetilde{\mathbf{{K}}}_{\mathcal{N}(i, w_2)}$ and $\widetilde{\mathbf{{V}}}_{\mathcal{N}(i, w_2)}$ respectively. 
\begin{align}
    \bar{\mathbf{K}}_i &= \text{Pooling}( \widetilde{\mathbf{{K}}}_{\mathcal{N}(i, w_2)}; \kappa, \xi) \\
    \bar{\mathbf{V}}_i &= \text{Pooling}( \widetilde{\mathbf{{V}}}_{\mathcal{N}(i, w_2)}; \kappa, \xi)
\label{eq:pool}
\end{align}
where $\kappa$ and $\xi$ are the pooling kernel size and stride size respectively. $\bar{\mathbf{K}}_i$ and $\bar{\mathbf{V}}_i$ are the compressed key, value matrices, and their size is $\xi$ times smaller than $\widetilde{\mathbf{K}}_i$ and $\widetilde{\mathbf{V}}_i$. 

The output of the second level pooling attention for token $i$ is calculated in Eqn. \ref{eq:second_att}. 
\begin{align}
\zvec_i^T = \text{Softmax}\left( \alpha \widetilde{\qvec}_i^T \bar{\mathbf{K}}_{i} \right) \bar{\mathbf{V}}^T_{i}
\label{eq:second_att}
\end{align}
In addition, we adopt a residual connection between the first level and second level attention modules, such that the final output of the two-level self-attention in Poolingformer is the sum of $\yvec_i$ (as in Eqn. \ref{eq:first_att}) and $\zvec_i$ (as in Eqn. \ref{eq:second_att}).



\textbf{Pooling:} we explore a few different pooling operations to compute $\bar{\mathbf{K}}_i$ and $\bar{\mathbf{V}}_i$ in our empirical studies, including the mean pooling, the max pooling and the convolution pooling ~\cite{DBLP:conf/iclr/WuFBDA19}. For a more comprehensive study, we introduce two trainable pooling mechanisms: the lightweight dynamic convolution pooling (LDConv) ~\cite{DBLP:conf/iclr/WuFBDA19} and its variant mean-LDConv:
The input matrix $\mathbf{V} = (\vvec_1,...,\vvec_m)$ is first chunked into a list of segments : $\left( (\vvec_1,...,\vvec_\kappa), (\vvec_{1+\xi},...,\vvec_{1+\xi+\kappa}), .. . \right)$ in the pooling according to the kernel size $\kappa$ and stride size $\xi$. The LDConv then maps each segment, i.e. $(\vvec_1,...,\vvec_\kappa)$, into a single vector for information compression in Eqn. \ref{eq:LDConv}
\begin{align}
\operatorname{LDConv}(\vvec_1,...,\vvec_\kappa)=\sum_{i=1}^{\kappa} \delta_{i} \cdot \vvec_i 
\label{eq:LDConv}
\end{align}
where $(\delta_1,...,\delta_\kappa)$ are called dynamic weights, computed by the context of $\vvec_{i=\lceil \frac{1+\kappa}{2} \rceil}$,
\begin{align}
    (\delta_1, ..., \delta_\kappa)^T = \text{Softmax}(\mathbf{W_{p}} \vvec_i)
\end{align}
$\mathbf{W_p} \in  \mathbb{R}^{\kappa \times d}$ is a learnable weight matrix.  
In the mean-LDConv, the dynamic weights $(\delta_1,...,\delta_\kappa)$ are computed by the mean of the context $\bar{\vvec} = \frac{1}{\kappa}\sum_{i=1}^{\kappa}{\vvec_i}$,
\begin{align}
    (\delta_1, ..., \delta_\kappa)^T = \text{Softmax}(\mathbf{W_{p}} \bar{\vvec})
\end{align}
A detailed comparison on different pooling approaches is presented in section \ref{subsubsec:ablation}.

\subsubsection{Task specific global attention}
\label{subsubsec:poolingfomer_global_att}
In some specific long document modeling tasks, i.e., Question Answering,  the question tokens are important to all the document tokens. Therefore, we follow Longformer~\cite{beltagy2020longformer} to append the indexes of query tokens into a global set $\mathcal{G} = \{ q_1, ..., q_l \}$ and allow all the tokens to attend to both the tokens in the global set and the tokens within its sliding window.

We integrate the global tokens into the first-level attention module in Poolingformer. The receptive field for each token $i$ (not in the global set) is the union of $\mathcal{N}(i, w_1)$ and $\mathcal{G}$. For the tokens in the global set, the receptive field is the entire text sequence.
\begin{equation}
\mathcal{N_G}(i, w_1) =
\begin{cases}
 \mathcal{N}(i, w_1) \cup \mathcal{G} & i \notin \mathcal{G} \\
 [\![1,...,n]\!] & i \in \mathcal{G}
\end{cases}
\end{equation}
The output for token $i$ of the first-level attention in Eqn. \ref{eq:first_att} is revised accordingly in Eqn. \ref{eq:first_att_global}
\begin{align}
\yvec_i^T = \text{Softmax}\left( \alpha \qvec_i^T \mathbf{K}_{\mathcal{N_G}{(i,w_1)}} \right) \mathbf{V}^T_{\mathcal{N_G}{(i,w_1)}}
\label{eq:first_att_global}
\end{align}

\subsubsection{Complexity Analysis}
\label{subsubsec:poolingfomer_complexity}
In this section, we simply analyze the complexity of Poolingformer. The computational complexity of the first-level sliding window attention is $\mathcal{O}(w_1 n)$. Considering $w_1$ is a constant and usually much smaller than $n$, the computational complexity can be simplified as $\mathcal{O}(n)$. The computational complexity of the second-level pooling attention is $\mathcal{O}(n\frac{w_2}{\xi})$, in which $w_2$ and $\xi$ are two hyper-parameters. Compared with $n$, we usually configure the ratio $\frac{w_2}{\xi}$ to be a relatively small constant. Therefore, the complexity of the second-level pooling attention is $\mathcal{O}(n)$. In summary, the overall complexity of Poolingformer is $\mathcal{O}(n)$, we list the computational complexity of different long document modeling methods in Table~\ref{table.complexity} for comparison.

\begin{table}[t]
\caption{Computational complexity of several related models.}
\begin{center}
\begin{tabular}{@{}lc@{}}
\toprule
Model & Complexity \\  \midrule
Transformer~\cite{vaswani2017attention} & $\mathcal{O}(n^2)$ \\
Reformer~\cite{KitaevKL20} &  $\mathcal{O}(n \log n)$\\
Cluster-Former~\cite{wang2020cluster} & $\mathcal{O}(n \log n)$ \\ 
Longformer~\cite{beltagy2020longformer} &   $\mathcal{O}(n)$ \\
BigBird~\cite{bigbird} &  $\mathcal{O}(n)$ \\
\hline
Poolingformer & $\mathcal{O}(n)$\\\bottomrule
\end{tabular}
\end{center}
\label{table.complexity}
\end{table}

\begin{table*}[ht]
\caption{Results on the dev set and the blind test set of NQ. We report the evaluation results of the precision (P), the recall (R), and the F1 score for both long-answer (LA) and short-answer (SA) tasks.}
\begin{center}
\resizebox{\linewidth}{!}{
\begin{tabular}{lcccccccccccc}
\toprule
\multicolumn{1}{l}{} & \multicolumn{3}{c}{NQ LA Dev} & \multicolumn{3}{c}{NQ LA Test} & \multicolumn{3}{c}{NQ SA Dev} & \multicolumn{3}{c}{NQ SA Test} \\ 
 & P & R & F1 & P & R & F1 & P & R & F1 & P & R & F1  \\
 \midrule
DocumentQA~\cite{clark2018simple} & 47.5 & 44.7 & \cellcolor[HTML]{ECF4FF}46.1 & 48.9 & 43.3 & \cellcolor[HTML]{ECF4FF} 45.7 & 38.6 & 33.2 & \cellcolor[HTML]{ECF4FF}35.7 & 40.6 & 31.0 & \cellcolor[HTML]{ECF4FF} 35.1  \\
DecAtt~\cite{parikh2016decomposable} + DocReader~\cite{chen2017reading} & 52.7 & 57.0 & \cellcolor[HTML]{ECF4FF}54.8 & 54.3 & 55.7 & \cellcolor[HTML]{ECF4FF}55.0 & 34.3 & 28.9 & \cellcolor[HTML]{ECF4FF}31.4 & 31.9 & 31.1 & \cellcolor[HTML]{ECF4FF}31.5 \\
BERTjoint~\cite{alberti2019bert} & 61.3 & 68.4 & \cellcolor[HTML]{ECF4FF}64.7 & 64.1 & 68.3 & \cellcolor[HTML]{ECF4FF}66.2 & 59.5 & 47.3 & \cellcolor[HTML]{ECF4FF}52.7 & 63.8 & 44.0 & \cellcolor[HTML]{ECF4FF}52.1\\
RikiNet~\cite{LiuGFYCJLD20} & 74.3 & 76.4 & \cellcolor[HTML]{ECF4FF}75.3 & - & - & \cellcolor[HTML]{ECF4FF} - & 61.4 & 57.3 & \cellcolor[HTML]{ECF4FF}59.3 & - & - & \cellcolor[HTML]{ECF4FF}- \\
\quad-Ensemble model& 73.3 & 78.7 & \cellcolor[HTML]{ECF4FF}75.9 & 78.1 & 74.2 & \cellcolor[HTML]{ECF4FF} 76.1 & 66.6 &  56.4 & \cellcolor[HTML]{ECF4FF}61.1 & 67.6 & 56.1 & \cellcolor[HTML]{ECF4FF}61.3 \\ 
ReflectionNet~\cite{wang2020no} & 79.4 & 72.7 & \cellcolor[HTML]{ECF4FF}75.9 & - & - & \cellcolor[HTML]{ECF4FF} - & 69.3 & 55.0 & \cellcolor[HTML]{ECF4FF}61.3 & - & - & \cellcolor[HTML]{ECF4FF}-\\
\quad-Ensemble model& 78.2 &  75.9 & \cellcolor[HTML]{ECF4FF} 77.0 & 76.8 & 77.6 & \cellcolor[HTML]{ECF4FF} 77.2 & 67.9 & 59.4  & \cellcolor[HTML]{ECF4FF}\textbf{63.4} & 70.4 & 58.8 & \cellcolor[HTML]{ECF4FF}\textbf{64.1} \\ 
\hline
Sparse Transformer~\cite{child2019generating} & - & - & \cellcolor[HTML]{ECF4FF}74.5 & - & - & \cellcolor[HTML]{ECF4FF}- & - & - & \cellcolor[HTML]{ECF4FF}56.1 & - & - & \cellcolor[HTML]{ECF4FF}-\\
Reformer~\cite{KitaevKL20} & - & - & \cellcolor[HTML]{ECF4FF}75.5 & - & - & \cellcolor[HTML]{ECF4FF}- & - & - & \cellcolor[HTML]{ECF4FF}56.4 & - & - & \cellcolor[HTML]{ECF4FF}- \\
BigBird-ETC~\cite{bigbird} & - & - & \cellcolor[HTML]{ECF4FF}- & 77.5 & 78.1 & \cellcolor[HTML]{ECF4FF}77.8 & - & - & \cellcolor[HTML]{ECF4FF}- & 63.7 & 53.4 & \cellcolor[HTML]{ECF4FF}57.9\\
Cluster-Former~\cite{wang2020cluster} & - & - & \cellcolor[HTML]{ECF4FF}76.5 & - & - & \cellcolor[HTML]{ECF4FF}- & - & - & \cellcolor[HTML]{ECF4FF}57.1 & - & - & \cellcolor[HTML]{ECF4FF}- \\
\quad-Ensemble model  & - & - & \cellcolor[HTML]{ECF4FF}- & 78.5 & 77.5 & \cellcolor[HTML]{ECF4FF}78.0 & - & - & \cellcolor[HTML]{ECF4FF}-  & 62.1 & 59.8 & \cellcolor[HTML]{ECF4FF} 60.9\\
\hline
Poolingformer  & 77.7 & 77.3 & \cellcolor[HTML]{ECF4FF}\textbf{77.5}  & - & - & \cellcolor[HTML]{ECF4FF}- & 62.3 & 55.3 & \cellcolor[HTML]{ECF4FF}58.6 & - & - & \cellcolor[HTML]{ECF4FF}- \\
\quad-Ensemble model  & - & - & \cellcolor[HTML]{ECF4FF}- & 78.5 & 81.2 & \cellcolor[HTML]{ECF4FF}\textbf{79.8} & - & - & \cellcolor[HTML]{ECF4FF}-  & 70.4 & 54.8 & \cellcolor[HTML]{ECF4FF} 61.6\\
\bottomrule
\end{tabular}
}
\end{center}
\label{table.NQMainResult}
\end{table*}
\section{Experiments}
\subsection{Datasets}
We evaluate Poolingformer on two long document tasks: Question Answering and Summarization. For QA, we  report the results on the monolingual Natural Question (NQ) and the multilingual TyDi QA. For long document summarization, we report the results on the arXiv dataset~\cite{cohan2018discourse}.

\textbf{Natural Questions:}
This dataset collected real questions in Google's search engine. Each question is paired with a Wikipedia page. Given a question and a document,  NQ requires the model to find (1) an answer span (short answer) and (2) a paragraph that contains the information required to answer the question (long answer). If the question can not be answered from the given document, the model is asked to return NULL ANSWER. NQ provides a blind test set consisting of 7,842 examples, whose labels are hidden to us. Any submission to the public leaderboard will be evaluated on this hidden dataset. The leaderboard system will produce the rank of the submission according to the F1 metric. 

\textbf{TyDi QA:}
TyDi QA is a multilingual question answering dataset consisting of 11 typologically diverse languages with 200K human-annotated question-answer pairs. Similar to NQ, each question is paired with a Wikipedia article. The model need to make two predictions: (1) index of the passage that answers the question (Passage Selection Task) (2) minimal span that completely answers the question (Minimal Answer Span Task). TyDi QA also provides a blind test set and maintains a leaderboard like NQ with the same evaluation metrics. 

\textbf{arXiv:}
arXiv~\cite{cohan2018discourse} is a long document summarization dataset collected from scientific repositories—arxiv.org. The dataset contains about 215k long Scientific papers and uses the paper abstract as the summary. About the length of the document, the mean, median and 90th percentile are about 5k, 6.1k and 16.5k, respectively. Following previous work, We use ROUGE-1, ROUGE-2, and ROUGE-L as automatic evaluation metrics. 

\subsection{Implementation Details}
\textbf{Question Answer:}

For NQ and TyDi QA , We split documents into multiple spans with a sliding window approach~\cite{alberti2019bert}. The size and stride of the sliding window are set to 4,096 and 1,568, respectively. Each instance is formed by a start placeholder, a question, and a document span. The question and the document span are separated by a special placeholder. Since many instances contain no answer, the number of negative instances and positive instances is imbalanced. We follow ~\citet{LiuGFYCJLD20} to sub-sample negative instances during training. The ratio of the sub-sampling set to 0.5. Similar to~\citet{alberti2019bert}, we use  token features to predict the short answer (Minimal Answer Span for TyDi QA). During inference, the distance between the start position and the end position is limited to 30 tokens. To predict Long Answer (Passage Selection for TyDi QA), we generate paragraph representations by applying a mean pooling to the tokens within the same paragraph. The answer type is predicted by the document representation which is the mean of all the paragraph representations. 

We use RoBERTa-large~\citep{liu2019roberta} for NQ and XLM-RoBERTa~\citep{ConneauKGCWGGOZ20} for TyDi QA to initialize our models for training Poolingformer. Both models contain 24 Transformer encoder Layers. 
Since the maximum length of our model is several times that of the pretrained model, we follow~\citet{beltagy2020longformer} to loop copying the position embedding of pretrained model to initialize our model. From the 15th to the 20th layer of our models, we apply two-level pooling self-attention, with other layers adopting the sliding window self-attention. The reason why we only utilize the two-level pooling attention in part of the layers is to avoid catastrophic forgetting of the prior knowledge in the initialization model. 
Since question tokens are very important in the QA tasks, we treat question tokens the global tokens, as described in~\ref{subsubsec:poolingfomer_global_att}. The window sizes of the first-level and second-level is set to 128 and 512, respectively. The pooling kernel size, stride size are set to 5, 4. We use Adam optimizer~\cite{adam} with linear learning rate decay. The batch size, the training epoch, the learning rate, and the learning rate warmup proportion are set to 64, 2, $2\times10^{-5}$ and 0.1 respectively. For the NQ leaderboard, the model we submitted is an ensemble of three models using different hyper-parameters. For the TyDi QA leaderboard, we use a single model for the submission.

\textbf{Summarization:}

For the arXiv dataset, we use the Encoder-Decoder framework following previous works~\cite{zhang2020pegasus,gidiotis2020divide}. 
Pretrained model BART~\cite{lewis2020bart} is used to initialize our model which consists of 12 encoder and 12 decoder layers. We expand the position embedding of encoder using the same method as QA. 
We apply the Poolingformer structure on the encoder side and keep the decoder structure unchanged. For encoder, the $6^{th}$ to $11^{th}$ layers adopt our two-level pooling self-attention and others adopt the single-level sliding window self-attention. Besides, we set the first token in encoder to global token 
as described in \ref{subsubsec:poolingfomer_global_att}. The sizes of the first-level and second-level window are set to 128 and 512 respectively. The pooling kernel size and stride size are set to 5 and 4 respectively. We use Adam optimizer~\cite{adam} with linear learning rate decay. The batch size, the training epoch, the learning rate, and the learning rate warmup step are set to 128, 10, $2\times10^{-4}$, 1000, respectively. During inference, the beam size and length penalty are set to 5, 2 respectively.

For all experiments, we use 8 NVIDIA Tesla V100 GPUs. All the experiments are conducted on Huggingface Transformers~\cite{wolf-etal-2020-transformers} and Fairseq~\cite{ott2019fairseq}. We utilize Gradient Checkpointing~\cite{chen2016training}, Apex\footnote{\url{https://github.com/NVIDIA/apex}}, and Gradient Accumulation to save GPU memory.
\subsection{Main Results}

\begin{table*}[ht]
\caption{Performance comparisons on the dev set and the blind test set of TyDi QA. We report the results using precision (P), recall (R), and F1 score for both the Passage Answer and the Minimal Answer tasks. }
\begin{center}
\resizebox{\linewidth}{!}{
\begin{tabular}{lcccccccccccc}
\toprule
\multicolumn{1}{l}{} & \multicolumn{3}{c}{Passage Answer Dev} & \multicolumn{3}{c}{Passage Answer Test} & \multicolumn{3}{c}{Minimal Answer Dev} & \multicolumn{3}{c}{Minimal Answer Test} \\ 
 & P & R & F1 & P & R & F1 & P & R & F1 & P & R & F1 \\
 \midrule
Tydiqa-baseline~\cite{clark2020tydi} &  63.1 & 57.0 & \cellcolor[HTML]{ECF4FF} 59.1 & 62.3  & 67.1 & \cellcolor[HTML]{ECF4FF} 64.4 &  41.3 & 35.3 & \cellcolor[HTML]{ECF4FF}50.5 & 56.4 & 50.1 & \cellcolor[HTML]{ECF4FF} 52.7\\
mBERT-mnlp & - & - & \cellcolor[HTML]{ECF4FF}- & 63.8 & 60.4 & \cellcolor[HTML]{ECF4FF}61.7 & - & - & \cellcolor[HTML]{ECF4FF}-  & 61.5 & 47.3 & \cellcolor[HTML]{ECF4FF} 53.2 \\
GAAMA (XLM-R)-with ARES system & - & - & \cellcolor[HTML]{ECF4FF}- & 73.6 & 72.1 & \cellcolor[HTML]{ECF4FF} 72.6 & - & - & \cellcolor[HTML]{ECF4FF}- & 70.8 & 62.2 & \cellcolor[HTML]{ECF4FF} 66.1 \\
BERT with language-clustered vocab~\cite{chung2020improving} & - & - & \cellcolor[HTML]{ECF4FF}78.0 & 77.4 & 78.0 & \cellcolor[HTML]{ECF4FF}77.7 & - & - & \cellcolor[HTML]{ECF4FF}65.4 & 67.2 & 60.2 & \cellcolor[HTML]{ECF4FF}63.4 \\
Poolingformer & \textbf{79.5} & \textbf{78.7} & \cellcolor[HTML]{ECF4FF}\textbf{79.1} & \textbf{80.4} & \textbf{78.8} & \cellcolor[HTML]{ECF4FF}\textbf{79.5} & \textbf{74.4} & \textbf{63.2} & \cellcolor[HTML]{ECF4FF}\textbf{68.5} &  \textbf{73.5} & \textbf{63.3} &\cellcolor[HTML]{ECF4FF}\textbf{67.7}  \\
 \hline
Lesser Human  & 84.4 & 74.5 & \cellcolor[HTML]{ECF4FF} 79.9 & - & - & \cellcolor[HTML]{ECF4FF}- & 70.8 & 62.4 & \cellcolor[HTML]{ECF4FF} 70.1 & - & - & \cellcolor[HTML]{ECF4FF}- \\ \bottomrule
\end{tabular}
}
\end{center}
\label{table.TydiqaMainResult}
\end{table*}

\subsubsection{Google NQ Results}

The results of both dev set and test set on NQ are shown in Table~\ref{table.NQMainResult}. The top block of the table shows the results of several approaches with input length of 512. The first three rows of the top block show the results of three multi-passage baseline models presented in the original NQ paper~\cite{kwiatkowski2019natural}. The fourth and fifth rows show two previous state-of-the-art models. 
RikiNet~\cite{LiuGFYCJLD20} adds Dynamic Paragraph Dual-Attention (DPDA) reader and multi-level cascaded answer predictor on top of the pretrained models.  ReflectionNet~\cite{wang2020no} is a two-phase model with an answer verification mechanism. These two models are proposed for NQ task and it is not easy to extend them to other tasks. The middle block lists the results of well-known and strong baselines designed for long documents, including Sparse Transformer~\cite{child2019generating}, Reformer~\cite{KitaevKL20}, Cluster-Former~\cite{wang2020cluster}, BigBird~\cite{bigbird}. The first three rows are borrowed from the Cluster-Former paper~\cite{wang2020cluster}. The bottom block shows the results from Poolingformer. It is clear that Poolingformer has a significant improvement over previous methods consistently in both dev set and test set. It is worth noting that PoolingFormer achieves the best result in LA task in both single model and ensemble model. For example, in the hidden LA test set, its improvement over the previous state of the art is 1.8\%(79.8 vs. 78.0). We treat this significant improvement, since NQ is an extremely competitive leaderboard and these scores are produced by a hidden dataset from the official NQ organizer.

\subsubsection{TyDi QA Results}

In Table~\ref{table.TydiqaMainResult}, we compare Poolingformer with Tydiqa-baseline~\cite{clark2020tydi} and previous state-of-the-art models. Tydiqa-baseline utilizes mBert~\cite{alberti2019bert} which is a multilingual extended version of Bert.~\citet{chung2020improving} improve multilingual models with language-clustered vocabularies. 
We show that Poolingformer achieves a significant improvement over previous state-of-the-art models , improved from 77.7 to 79.5 in the Passage Answer task and 66.1 to 67.7 in theMinimal Answer tasks. 
The bottom block is a lesser estimate of human performance from~\citet{clark2020tydi}. Poolingformer further narrows the gap between machine and human performance. Without the ensemble approach, the gap between Poolingformer and human performance is only 0.4\% and 2.3\%.  At the time of our submission (25 Jan. 2021), Poolingformer achieves the new state-of-the-art result on both LA (F1 79.5) and SA (F1 67.7) on the TyDi QA leaderboard. All of these results demonstrate that Poolingformer is simultaneously shining in multilingual comprehension tasks.

\begin{table}[ht]
\caption{ The results on the arXiv test set. We report the results of ROUGE-1 (R-1), ROUGE-2 (R-2), and ROUGE-L (R-L). }
\begin{center}
\resizebox{\linewidth}{!}{
\begin{tabular}{@{}lccc@{}}
\toprule
Model  & R-1 & R-2 & R-L \\ \hline
Sent-PTR~\cite{pilault2020extractive} &  42.32 & 15.63 & 38.06 \\
Extr-Abst-TLM~\cite{pilault2020extractive}  &  41.62 & 14.69 & 38.03  \\
PEGASUS~\cite{zhang2020pegasus} & 44.21 & 16.95 & 38.83  \\
Dancer~\cite{gidiotis2020divide} & 45.01 & 17.60 & 40.56  \\
\hline
BigBird~\cite{bigbird} & 46.63 & 19.02 & 41.77  \\
LED$_{4k}$~\cite{beltagy2020longformer} & 44.40 & 17.94 & 39.76  \\
LED$_{16k}$~\cite{beltagy2020longformer} & 46.63 & 19.62 & 41.83  \\
\hline
Poolingformer$_{4k}$ & 47.86 & 19.54 & 42.35 \\
Poolingformer$_{16k}$ & \textbf{48.47} & \textbf{20.23} & \textbf{42.69} \\
\bottomrule
\end{tabular}
}
\end{center}
\label{table.arxivResult}
\end{table}
\subsubsection{Summarization Results}
The result on the arXiv test set is shown in table ~\ref{table.arxivResult}. The top block presents previous state-of-the-art methods with shorter input sequences. Sent-PTR~\cite{pilault2020extractive} is an extractive model that uses hierarchical LSTMs and a sentence pointer to select key sentences as the summary. Extr-Abst-TLM~\cite{pilault2020extractive} is a two-phase model that generates summaries based on sentences selected by an extractive model. PEGASUS~\cite{zhang2020pegasus} is a large pretrained model specifically for abstractive summarization with an input length up to 1,024. Dancer~\cite{gidiotis2020divide} breaks a long document into multiple sections to produce partial summaries for different sections and then produces a final complete summary based on the partial summaries. 

The middle and bottom blocks show the results of several long document modeling methods for longer input sequence. Both BigBird and LED (Longformer-Encoder-Decoder) use both the slide local attention and the global attention mechanism to encode long documents. BigBird~\cite{bigbird} initializes and continuously pretrain the model with PEGASUS which is dedicated to the summarization task. Following LED~\cite{beltagy2020longformer}, Poolingformer is initialized from BART~\cite{lewis2020bart} without the continuous pretraining process.  We evaluate the performance of Poolingformer with input lengths of both 4K and 16K. One can clearly see that Poolingformer with input length of 16k greatly outperforms previous state-of-the-art models. Even if the input length is reduced to 4k, Poolingformer can still achieve the best on ROUGE-1 and ROUGE-L.

In addition, Poolingformer achieves a better computational complexity than models with single-level local attention. On this dataset, LED~\cite{beltagy2020longformer} sets the local attention one-side window size to 512 to increase the model's receptive fields. That means the complexity of LED is $\mathcal{O}(1024 \times n)$. With the same receptive field, the complexity of Poolingformer's two-level attention is $\mathcal{O}((256+1024/4)\times n)$, which accounts for only half complexity of LED. In other words,  Poolingformer can greatly outperform LED in both accuracy and complexity.  This demonstrates the effectiveness of the two-level pooling attention schema from both dimensions. 


\subsection{Ablation Study}
\label{subsubsec:ablation}

For the sake of saving computational resources, we conduct all the ablation studies using one-fifth of the NQ training set using the base-size model. This model is initialized from RoBERTa-base. The $6^{th}-8^{th}$ layers of the model adopt two-level pooling self-attention, and other layers adopt sliding window self-attention. The sizes of the first-level, second-level window, the pooling kernel size and stride size are set to 128, 512, 5 and 4 respectively. 

\textbf{Performance improvements of long document modeling}:
The top block of Table~\ref{table.ablationWindowSize} shows a simple setting without the pooling attention. We first explored the advantages of long context modeling. Following previous work, we evaluate the RoBERTa model with the input length of 512.  We observe that other models supporting longer input length consistently produce better results than RoBERTa on the LA task .

\begin{table}[t]
\caption{Ablation study of Poolingformer$_{base}$ with different window lengths on NQ dev set. $w_1$: the size of the first level window. $w_2$: the size of the second level window. $C$: the compression rate of the second level window controlled by adjusting the kernel size and stride size of the pooling.}
\begin{center}
\resizebox{\linewidth}{!}{
\begin{tabular}{@{}cccccc@{}}
\toprule
Setting  & $w_1$ & $w_2$ & $C$ & LA F1 & SA F1 \\ \hline
RoBERTa$_{base}$ & - & - & - & 63.8 & 43.2  \\
Poolingformer$_{base}$ & 128 & - & - & 66.3 & 43.1  \\
Poolingformer$_{base}$ & 256 & - & - & 67.4 & 43.4 \\
Poolingformer$_{base}$ & 512 & - & - & 66.1 & 42.6 \\
\hline
Poolingformer$_{base}$ & 128 & 256 & 4 & 67.9 & 45.0 \\
Poolingformer$_{base}$ & 128 & 512 & 4 & \textbf{68.7} & \textbf{45.2} \\
Poolingformer$_{base}$ & 128 & 2,048 & 8 & 66.9 & 42.6 \\
Poolingformer$_{base}$ & 128 & 2,048 & 16 & 67.0 & 44.4 \\
\bottomrule
\end{tabular}
}
\end{center}
\label{table.ablationWindowSize}
\end{table}
\textbf{Useful but redundant information from distant tokens:} 
From the second to the fourth rows in Table~\ref{table.ablationWindowSize}, 
we remove the second-level window and explore the relationship between the size of the first-level window and task performance. We may expect the performance becomes better for a larger window size. But the results show that it achieves the best performance when the sliding window size is set to 256. We conjecture that the reason for the poor performance of 512 windows size is that the self-attention mechanism is difficult to deal with remote token due to redundancy noise. 
In the bottom two rows of the Table~\ref{table.ablationWindowSize}, the second-level window size is set to the entire input sequence. We compress the sequence length by $C$ times by adjusting the kernel and stride size in pooling attention. Each token will attend to the tokens in the first-level window and tokens compressed from the entire sequence. From the results, we can see this approach does not work very well. We think that for every distant token, there may be too little useful information to compute attention. With these findings, we designed a two-level pooling attention mechanism to perform coarse-grained compression on farther tokens. For the tokens that are very far away, we will discard them directly. 
In this way, tokens can pay more attention to key information and reduce computation and memory consumed.

\begin{table}[ht]
\caption{Ablation study of pooling and fusion approaches.}
\begin{center}
\resizebox{\linewidth}{!}{
\begin{tabular}{@{}lcc@{}}
\toprule
 Setting & LA F1 & SA F1 \\ \midrule
Poolingformer$_{base}$(Without $2nd$ level window) & 66.3 & 43.1 \\ \midrule
Poolingformer$_{base}$(MEAN) & 68.5 & 43.7  \\
Poolingformer$_{base}$(MAX) & 68.6 & \textbf{45.3}  \\ 
Poolingformer$_{base}$(LDConv) & \textbf{68.7} & 45.2  \\ 
Poolingformer$_{base}$(MeanLDConv) & 67.7 & 44.1 \\ 
\midrule
Poolingformer$_{base}$(LDConv, \textit{Mix}) & 67.5 & 44.6  \\ 
Poolingformer$_{base}$(LDConv, \textit{Weight Sharing}) & 67.2 & 44.2  \\ 
\bottomrule
\end{tabular}
}
\end{center}
\label{table.ablationPoolingWay}
\end{table}

\textbf{Impact of different pooling and fusion approaches:} 
In the experiment, we have explored four different pooling methods while keeping other settings unchanged. The results are shown in Table~\ref{table.ablationPoolingWay}. MEAN and MAX represent Mean pooling and Max pooling, respectively. LDConv refers to stride lightweight and dynamic convolution~\cite{DBLP:conf/iclr/WuFBDA19}, as we discussed in Eqn. \ref{eq:LDConv} . Mean-LDConv is an variant of LDConv, refers to the weighted sum of token embeddings within the pooling window, where the weight is dynamically generated using the mean and linear layer. The detail of LDConv and Mean-LDConv is given in section~\ref{subsubsection:poolingattention}. As presented in Table~\ref{table.ablationPoolingWay}, 
LDConv and MAX are slightly better than others. We defer a more comprehensive study of different pooling approach in future work. 


We explore another two different settings of poolingformer: \textit{Mix} and \textit{Weight Sharing}. In~\textit{Mix}, the second-level pooling attention module is built upon the input embeddings instead of the output of the first-level attention. To be more clear, it replaces $\mathbf{Y}$ with $\mathbf{X}$ in Eqn. \ref{eq:qkv_2} in ~\textit{Mix} setting. From Table ~\ref{table.ablationPoolingWay}, we can see that the Poolingformer in the \textit{Mix} setting performs worse on NQ tasks, which illustrates the 
effectiveness of stacking two level attentions in Poolingformer.   
In~\textit{Weight sharing}, the first level and second level share the linear mapping matrices $\mathbf{W_q}$, $\mathbf{W_k}$, and $\mathbf{W_v}$ in Eqn.\ref{eq:qkv} and Eqn.\ref{eq:qkv_2}. From Table~\ref{table.ablationPoolingWay}, we observe that the default setting produces better performance.  

\begin{table}[t]
\caption{Ablation study of the number of Poolingformer layer.}
\begin{center}
\begin{tabular}{@{}lcc@{}}
\toprule
 Setting & LA F1 & SA F1 \\ \midrule
Poolingformer$_{base}$(0 layers) & 66.3 & 43.1 \\ \midrule
Poolingformer$_{base}$(1 layers) & 68.0 & 44.5  \\
Poolingformer$_{base}$(3 layers) & \textbf{68.7} & \textbf{45.2}  \\
Poolingformer$_{base}$(6 layers) & 67.5 & 43.7  \\ 
Poolingformer$_{base}$(all layers) & 65.0 & 41.5 \\ \bottomrule
\end{tabular}
\end{center}
\label{table.ablationPoolinglayer}
\end{table}

\textbf{Impact of Poolingformer layer number:} 
As shown in Table~\ref{table.ablationPoolinglayer}, an appropriate number of Poolingformer layers can greatly improve the model performance, up to 2.4 points and 2.1 points in terms of LA F1 and SA F1, respectively.  This continues to demonstrate the value of the Poolingformer layers. On the other hand, additional Poolingformer layers do not always lead to a better the performance. We observe some performance degradation when all the layers are replaced with the Poolingformer layers. Although the Poolingformer layer can effectively make use of distant information, it is still not fully compatible with the existing pretrained models. This may lead to some catastrophic forgetting of the information in the pretrained models. It is actually a trade-off between distant information and the prior knowledge of the pretrained models. Our experience shared with us that the best results often happen when the number of Poolingformer layer is one fourth of the total number of layers.

\section{Related Work}

The core limitation of Transformer in long document modeling is the computational complexity since the self-attention mechanism can grow quadratically to the sequence length. There are two widely adopted approaches to mitigate this problem.  One is to use kernel functions, random projection, and others to approximate or eliminate the dot product in self-attention.  Synthesizer~\cite{tay2020synthesizer} directly uses trainable parameters to generate attention weights, avoiding the dot-product interactions. Performer~\cite{choromanski2020rethinking} and Linear Transformer~\cite{katharopoulos2020transformers} view the attention mechanism through kernelization and design different kernel functions to approximate the attention matrix. Compared with the original attention, these methods can reduce the complexity to linearity. But the performance of these methods comes with no theoretical guarantee. Moreover, it is difficult to make them compatible with existing pretrained models.

Another method is sparse attention, which focuses on making each token attend to less but more important context. Generally, the most important context is the local context. One simple way is the blockwise pattern~\cite{qiu2020blockwise}, which cuts the input sequence into multiple fixed chunks, and each token only attends to its neighbors within the same chunk. Furthermore, BP-Transformer~\cite{ye2019bp} uses the binary partitioning tree to hierarchically block the sequence, and each token receives information from different blocks according to distance. Another approach is the sliding window attention pattern that each token can attend to the neighbors in a sliding window. Longformer~\cite{beltagy2020longformer}, BigBird~\cite{bigbird} use this attention pattern to capture local information, and use global attention to capture global information which is similar to Star Transformer~\cite{guo2019star}. Moreover, Sparse Transformer~\cite{child2019generating} and Longformer~\cite{beltagy2020longformer} propose dilated window attention pattern which is similar to dilated convolution~\cite{DilatedConvolutions}. This pattern works well in autoregressive language modeling, but it is also not compatible with existing pretrained models. Linformer~\cite{wang2020linformer} assumes that the attention mechanism matrix is low-rank, and utilizes linear mapping to compress sentence sequences. Another related work is Memory Compressed Attention~\cite{LiuSPGSKS18}, which adopts stride convolution to compress sentence information in the decoder and its computational complexity does not increase linearly with length. Cluster-Former~\cite{wang2020cluster}, Reformer~\cite{KitaevKL20}, and Routing Transformer~\cite{roy2020efficient} utilize locally sensitive hashing and clustering methods to assign tokens with high similarity into buckets. Each token only attends to the tokens within its bucket. 

\section{Conclusion}
We introduce Poolingformer, a two-level attention model for long sequence modeling with linear complexity. In the first level attention, it uses a smaller sliding window pattern to aggregate information from neighbor tokens. In the second level attention, it increases the receptive fields with a larger window size, followed by a pooling operation on both the key and value vectors to reduce the computational cost. Poolingformer achieves new state-of-the-art performance on long-document QA tasks and shows superior performance on long-document summarization task.  For future work, we will continue to explore continuous improvement of Poolingformer from the following perspectives: 1) Theoretical analysis of the proposed multi-level attention in contrast to the classical single-level self-attention. 2) Extend Poolingformer to other types of long sequence data, such as image and music.  

\section{Acknowledgement}
We would like to thank Dayiheng Liu, Weizhen Qi for helpful discussions.

\newpage

\bibliography{main}
\bibliographystyle{icml2021}
\end{document}